\documentclass[journal]{IEEEtran}%duoble
\newcommand{\ignore}[1]{}
\usepackage[utf8]{inputenc}
\usepackage[T1]{fontenc}
\usepackage{amsmath,epsfig}
\usepackage{graphicx}
\usepackage{bm}
\usepackage{amssymb}
\usepackage{cite}
\usepackage{array}
\usepackage{makecell}
\usepackage{extarrows}
\usepackage{url}
\usepackage{float}
\usepackage{pifont}
\usepackage{color}

\usepackage{multirow}
\usepackage{booktabs}
\usepackage{blkarray}
\usepackage{multirow}
\usepackage{array}
\usepackage{color}
\usepackage{pmat}
\begin{document}
%
% paper title
% can use linebreaks \\ within to get better formatting as desired
\title{Unity is Strength: Unifying Convolutional and Transformeral Features for Better Person Re-Identification}
\author{Yuhao~Wang,
        Pingping~Zhang$^{*}$, \emph{IEEE Member},
        Xuehu~Liu,
        Zhengzheng Tu
        and~Huchuan~Lu, \emph{IEEE Fellow}

\thanks{
Copyright (c) 2024 IEEE. Personal use of this material is permitted. However, permission to use this material for any other purposes must  be obtained from the IEEE by sending an email to \textcolor{blue}{\underline{pubs-permissions@ieee.org}}.

%This work was supported in part by the National Natural Science Foundation of China (No.62101092), the Open Project of Anhui Provincial Key Laboratory of Multimodal Cognitive Computation, Anhui University (No.MMC202102) and the Fundamental Research Funds for the Central Universities (No.DUT23BK05).
($^{*}$ Corresponding author: Pingping Zhang.)

YH. Wang and PP. Zhang are with School of Future Technology, School of Artificial Intelligence, Dalian University of Technology, Dalian, 116024, China. (Email: 924973292@mail.dlut.edu.cn; zhpp@dlut.edu.cn)

XH. Liu is with School of Computer Science and Artificial Intelligence, Wuhan University of Technology, Wuhan, 430070, China. (Email: liuxuehu@whut.edu.cn)

ZZ. Tu is with School of Computer Science and Technology, Anhui University, Hefei, 230039, China. (Email: zhengzhengahu@163.com)

HC. Lu is with School of Information and Communication Engineering, Dalian University of Technology, Dalian, 116024, China.  (Email: lhchuan@dlut.edu.cn)
}
}
% make the title area
\maketitle
% The paper headers
\markboth{IEEE Transactions on Intelligent Transportation Systems}{}
\begin{abstract}
Person Re-identification (ReID) aims to retrieve the specific person across non-overlapping cameras, which greatly helps intelligent transportation systems.
As we all know, Convolutional Neural Networks (CNNs) and Transformers have the unique strengths to extract local and global features, respectively.
Considering this fact, we focus on the mutual fusion between them to learn more comprehensive representations for persons.
In particular, we utilize the complementary integration of deep features from different model structures.
We propose a novel fusion framework called FusionReID to unify the strengths of CNNs and Transformers for image-based person ReID.
More specifically, we first deploy a Dual-branch Feature Extraction (DFE) to extract features through CNNs and Transformers from a single image.
Moreover, we design a novel Dual-attention Mutual Fusion (DMF) to achieve sufficient feature fusions.
The DMF comprises Local Refinement Units (LRU) and Heterogenous Transmission Modules (HTM).
LRU utilizes depth-separable convolutions to align deep features in channel dimensions and spatial sizes.
HTM consists of a Shared Encoding Unit (SEU) and two Mutual Fusion Units (MFU).
Through the continuous stacking of HTM, deep features after LRU are repeatedly utilized to generate more discriminative features.
Extensive experiments on three public ReID benchmarks demonstrate that our method can attain superior performances than most state-of-the-arts.
{The source code is available at https://github.com/924973292/FusionReID.}
\end{abstract}
\IEEEpeerreviewmaketitle
\section{Introduction}
\iffalse
%-------------------------------------------------------------
\fi
\IEEEPARstart{P}erson Re-identification (ReID) aims to retrieve the same person across different scenes and camera views.
\begin{figure}[htpb]
\centering
\includegraphics[width=0.42\textwidth]{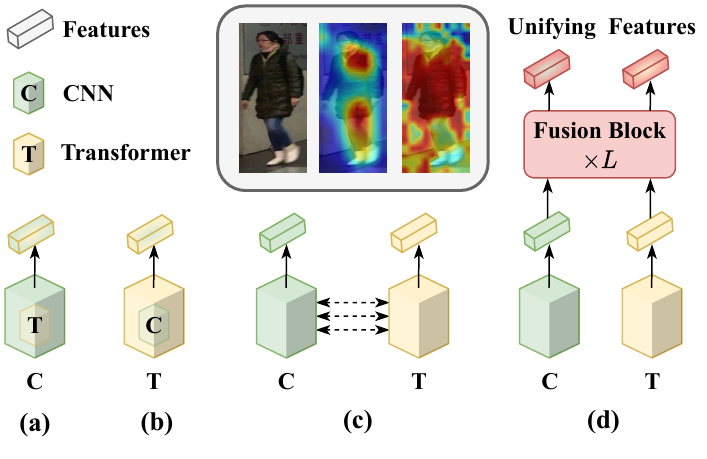}
\vspace{-2mm}
\caption{Different unities of CNNs and Transformers. (a) Transformer is embedded into CNN; (b) CNN is embedded into Transformer; (c) Parallel interaction between CNNs and Transformers; (d) Our framework focuses on the deep feature fusion.}
\label{fig1}
\vspace{-6mm}
\end{figure}
With the advancements in autonomous driving and intelligent surveillance systems, person ReID gains significant attention~\cite{bai2020deep}.
Due to variations in viewpoints, lighting conditions and postures, the appearance of the same person can be significantly different~\cite{yang2021learning}.
Therefore, most existing methods focus on learning robust and discriminative features to improve the performance.
Currently, there are two kinds of outstanding methods for extracting person features, of which one based on Convolutional Neural Networks (CNNs) and the other based on Transformers~\cite{vaswani2017attention}.
In fact, each kind of methods offers its own strengths and limitations.
As shown in the middle of Fig.~\ref{fig1}, the three images from left to right are the original image, the feature map from ResNet50~\cite{he2016deep}, and the feature map from Vision Transformer (ViT)~\cite{dosovitskiy2020image}.
{One can observe that CNN-based methods~\cite{chen2024visible,yin2024robust,shi2023dual,shi2024multi,chen2019abd,dai2019batch,wang2020high,zhang2021hat,zhang2020relation,wang2022nformer} tend to extract local representations of person appearances, while lacking a comprehensive perspective.
In contrast, Transformer-based methods~\cite{he2021transreid,zhu2021aaformer,zhang2024magic,wang2024top,wang2024other,yu2024tf,liu2024video,liu2023deeply,lu2023learning} concentrate on establishing dependencies between image patches and capture global representations in terms of structural information, such as poses, body shapes and contours.}
Nevertheless, they may struggle to extract fine-grained features.
Given the aforementioned facts, it is imperative to leverage the strengths of both CNNs and Transformers to enhance the ability of feature representations.
As shown in Fig.~\ref{fig1}(a), some methods~\cite{zhang2021hat} enhance CNNs by exploiting non-local blocks and self-attention mechanisms~\cite{vaswani2017attention}.
As shown in Fig.~\ref{fig1}(b), some methods~\cite{wu2021cvt} leverage convolutional layers to improve the local capacity of Transformers.
As shown in Fig.~\ref{fig1}(c), some methods\cite{chen2022mobile} build the interaction  between CNNs and Transformers with intermediate features.
Different from previous methods, as shown in Fig.~\ref{fig1}(d), we directly fuse deep features from CNNs and Transformers.
Thus, in this paper, we propose a novel fusion framework called FusionReID, which unifies the strengths of CNNs and Transformers for effective person ReID.
Our framework comprises of two main components: a Dual-branch Feature Extraction (DFE) and a Dual-attention Mutual Fusion (DMF).
More specifically, our DFE includes a CNN branch and a Transformer branch, and enables to independently extract two types of feature maps from a single image.
Then, we pass the features into our DMF.
The DMF comprises Local Refinement Units (LRU) and Heterogenous Transmission Modules (HTM).  In fact, each HTM is composed of a Shared Encoding Unit (SEU) and two Mutual Fusion Units (MFU).
At the beginning, we utilize LRU to align the channel dimensions and spatial sizes between two types of deep features.
Then, SEU and MFU are deployed to achieve shared feature encoding and mutual feature fusion, respectively.
Within the SEU, self-attention is utilized to capture the dependencies between local features.
In contrast, MFU employs a cross-attention to explore the associations between local and global features across two branches.
Through the stacking of HTM, it continuously fuses deep features from different branches, leading to the effective unity of CNNs and Transformers.
We conduct comprehensive experiments on three large-scale person ReID benchmarks.
Experimental results demonstrate that our method could attain superior performance than most state-of-the-arts.
In short, our contributions can be summarized as follows:
\begin{itemize}
\item
We propose a new fusion framework called FusionReID to unify the strengths of CNNs and Transformers for image-based person ReID.
\item
We design a novel Dual-attention Mutual Fusion (DMF), which can generate more discriminative features with stacking Heterogenous Transmission Modules (HTM).
\item
Our proposed framework achieves superior performances than most state-of-the-art methods on three public person ReID benchmarks.
\end{itemize}

The rest of this work is organized as follows:
Sec.~II describes the related works.
Sec.~III presents our approach, including the introduction of DFE, DMF and the objective function.
Sec.~IV elaborates the datasets, experimental settings, experimental results and visualization analysis.
Sec.~V presents the conclusion and future work.
%-------------------------------------------------------------------------
\section{Related Work}
\subsection{CNN-based Person Re-identification}
In recent years, person ReID has achieved impressive performances with deep learning.
Typically, image-based person ReID approaches focus on extracting discriminative features~\cite{zhang2021hat}.
Prior to Transformers, CNN-based methods have dominated person ReID.
The importance of local features is first suggested in~\cite{yi2014deep}.
Typical methods (\emph{e.g.}, PCB~\cite{sun2018beyond}, MGN~\cite{wang2018learning}) divide person images into several stripes to obtain multi-grained representations.
Besides, Yao~\emph{et al.}~\cite{yao2019deep} introduce a part-based loss to extract fine-grained features from different local parts.
Auxiliary information such as human pose~\cite{sarfraz2018pose,zheng2019pose}, parsing~\cite{he2019foreground,zhang2019densely} is also used for discriminative feature extraction.
Other than focusing on the local parts of persons, other researchers try to utilize CNNs to extract more contextual features~\cite{luo2019bag,xiao2016learning}.
However, limited by the local receptive field of CNNs, the contextual relationship between the human parts is not fully extracted.
Inherently, the attention mechanism in Transformers adaptively models the relationship between all the human parts.
Thus, the superiority of Transformers can help us alleviate this problem.
%------------------------------------------------------------------------
\subsection{Transformer-based Person Re-identification}
Transformers have also been introduced to person ReID.
As a primary work, He~\emph{et al.}~\cite{he2021transreid} present a pure Transformer-based method named TransReID for object re-identification.
Zhu~\emph{et al.}~\cite{zhu2021aaformer} adopt an automatic alignment to extract local features for accurate person ReID.
Li~\emph{et al.}~\cite{li2018harmonious} propose a harmonious attention network to learn robust feature representations.
Li~\emph{et al.}~\cite{li2021diverse} introduce an encoder-decoder Transformer architecture for person ReID.
Li~\emph{et al.}~\cite{li2025adaptive} leverage frequency-aware information for more robust feature extraction.
Although these methods show better results, Transformers lack the inductive bias of CNNs.
It has a weak perception of details such as textures, which leads to the ambiguity in persons and backgrounds.
Fortunately, CNNs are sensitive to low-level information, so it is necessary to introduce the desired properties for person ReID.
\begin{figure*}[th]
\centering
\includegraphics[width=0.85\textwidth]{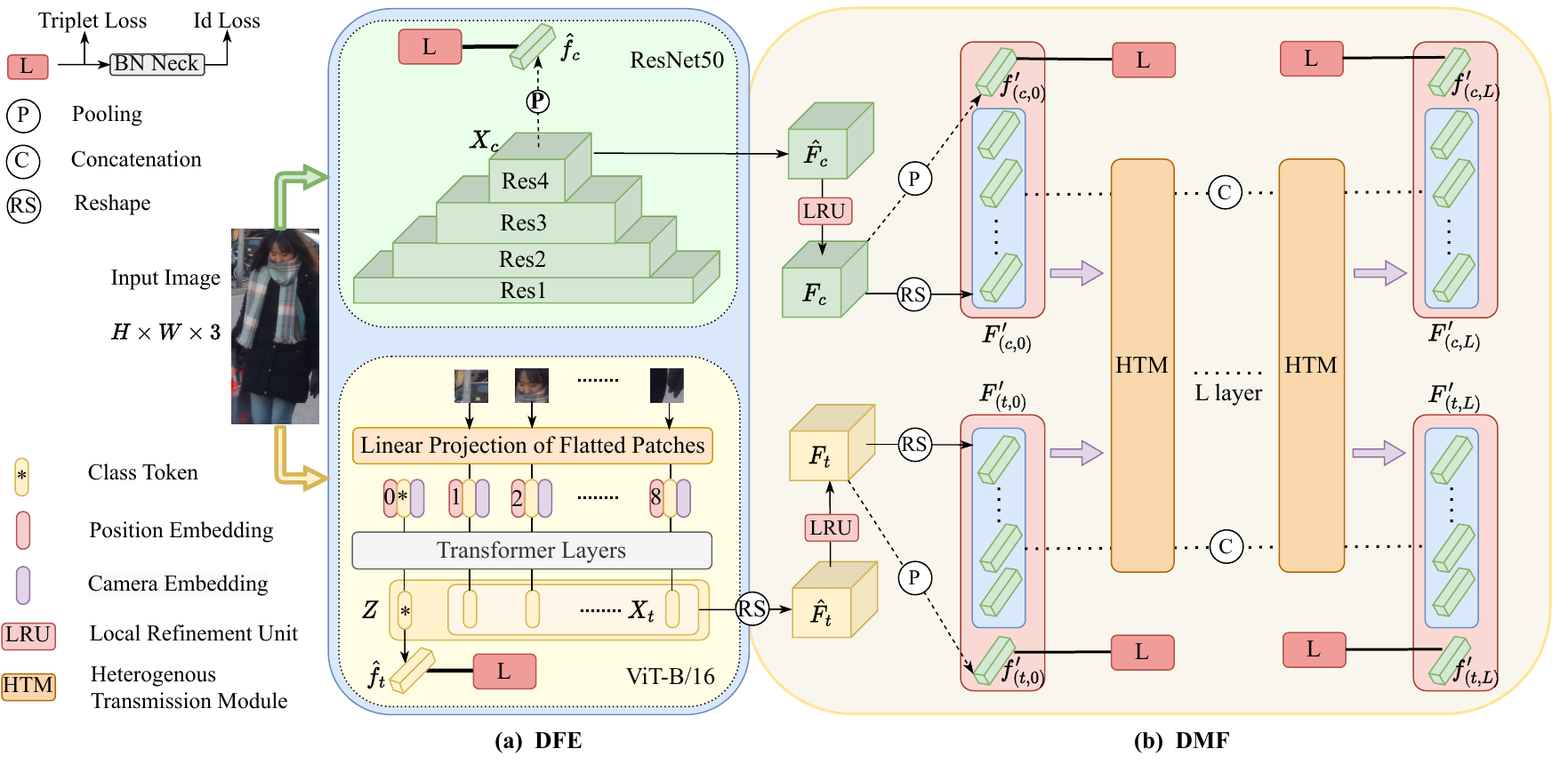}
\vspace{-2mm}
\caption{Overall architecture of the proposed FusionReID. With CNN and Transformer backbones, the Dual-branch Feature Extraction (DFE) is used to extract two types of deep features from the same image. Afterwards, the Dual-attention Mutual Fusion (DMF) is adopted to fuse deep features. More specifically, Local Refinement Units (LRU) are utilized to adjust the channel dimensions and spatial sizes. Heterogenous Transmission Modules (HTM) are stacked to enhance the heterogenous features. The framework allows the combination of different backbones, resulting in highly discriminative features.}
\label{fig2}
\vspace{-2mm}
\end{figure*}
\subsection{Combination of CNNs and Transformers}
Some researchers are focusing on how to effectively fuse the features of CNNs and Transformers.
For example, Conformer~\cite{gulati2020conformer} and Mobile-Former~\cite{chen2022mobile} are designed with a parallel structure that allows continuous interaction of deep features.
Dai~\emph{et al.}~\cite{dai2021coatnet} unify depthwise convolutions and self-attentions for visual recognition.
Zhang~\emph{et al.}~\cite{zhang2021hat} insert Transformer layers into different levels of features in CNNs.
Chen~\emph{et al.}~\cite{chen2019abd} enhance the features through channel and spatial attention modules for CNN backbones.
For robust person ReID, Wang~\emph{et al.}~\cite{wang2022nformer} utilize a CNN backbone and the neighbor cluster method for Transformers.
These previous methods~\cite{10196489,yan20233d,xie2024ghostformer} typically insert one structure into another or interactively stack two structures.
In the case of two-branch structures, the continuous interaction starts at low-level feature extraction, potentially leading to the destruction of deep features.
In this work, we abandon the complex structures and use a parallel feature fusion, preserving the inherent properties of CNNs and Transformers.
Besides, our method is general and the used backbone can be easily replaced.
%-----------------------------------------------
\section{Proposed Method}
As shown in Fig. \ref{fig2}, we propose a novel framework named FusionReID to unify convolutional and transformeral features for person ReID.
It consists a Dual-branch Feature Extraction (DFE) and a Dual-attention Mutual Fusion (DMF).
In DFE, we first utilize the ResNet50~\cite{he2016deep} and ViT-B/16~\cite{dosovitskiy2020image} as our backbones to extract convolutional features and transformeral features, respectively.
Afterwards, these features are passed into DMF to achieve the deep fusion.
The DMF consists of Local Refinement Units (LRU) and Heterogenous Transmission Modules (HTM).
LRU is used for feature dimension alignment. Meanwhile, HTM includes a Shared Encoding Unit (SEU) and two Mutual Fusion Units (MFU), which achieve feature shared encoding and deep mutual fusion.
To enhance the heterogenous features, HTM is continuously stacked.
During training, we deploy the cross-entropy loss~\cite{szegedy2016rethinking} and the triplet loss~\cite{hermans2017defense} to supervise the whole framework.
\subsection{Dual-branch Feature Extraction}
To jointly extract deep features, previous works mainly modify the basic structures of CNNs and Transformers.
Different from them, we directly utilize the CNNs and Transformers as our backbone networks.
As shown in the left part of Fig.~\ref{fig2}, the ResNet50~\cite{he2016deep} is deployed to extract the convolutional features $\textbf{X}_c \in R^{D_c\times H_c \times W_c}$.
Here $D_c$, $H_c$ and $W_c$ present the channel number, height and weight, respectively.
Meanwhile, the ViT-B/16~\cite{dosovitskiy2020image} is used to extract transformeral features $\textbf{Z} \in R^{D_t\times (1+N)}$.
Here, $D_t$ denotes the embedding dimensions and $N$ represents the number of image patches. Additional learnable class token is also utilized to aggregate the patch features.
We also add the camera embedding~\cite{he2021transreid} to all the tokens to enhance the feature discrimination.
In the CNN branch, $\textbf{X}_c$ is followed by a Generalized Mean Pooling (GeMP)~\cite{radenovic2018fine} to obtain the final feature vector $\hat{f}_c \in R^{D_c}$.
In the Transformer branch, $\textbf{Z}$ is divided into $\hat{f}_t \in R^{D_t}$ and $\textbf{X}_t \in R^{D_t\times N}$.
Furthermore, $\hat{f}_c$ and $\hat{f}_t$ are utilized for loss supervision.
$\textbf{X}_c$ and $\textbf{X}_t$ are passed into DMF for the subsequent feature fusion.
\begin{figure*}[htpb]
\centering
\includegraphics[width=0.85\textwidth]{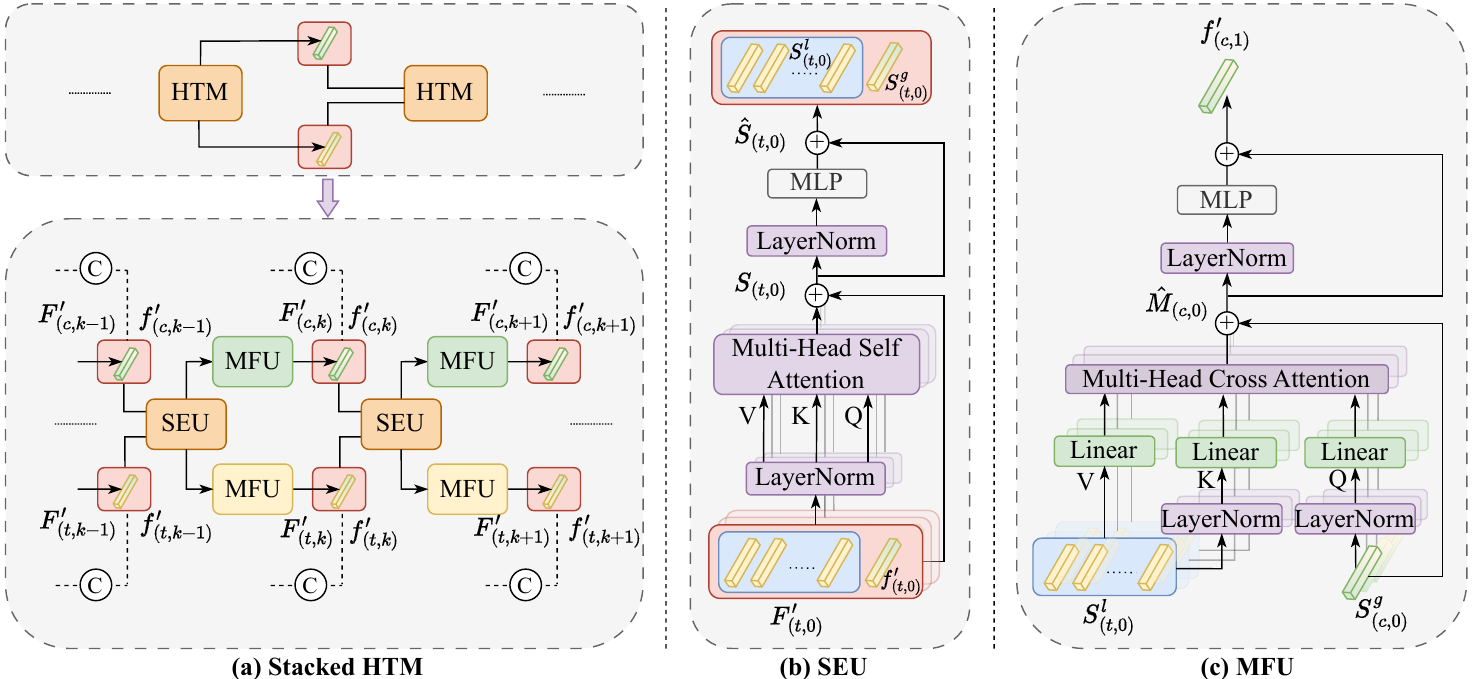}
\vspace{-2mm}
\caption{Illustrations of our key modules. (a) Details of continuously stacked HTM; (b) SEU for deep feature enhancement; (c) MFU for deep feature fusion. We take the Transformer branch as an example. The CNN branch has a similar structure.}
\label{fig3}
\vspace{-4mm}
\end{figure*}
\subsection{Dual-attention Mutual Fusion}
After extracting features by DFE, we propose a novel DMF for deep feature fusion, as depicted in Fig.~\ref{fig2}.
Although feature concatenation can be utilized, this simple fusion strategy fails to fully exploit local features from CNNs and global features from Transformers.
Therefore, our approach adopts the attention mechanism, including self-attention and cross-attention, to achieve a deep mutual fusion between the convolutional and transformeral features.
More specifically, our proposed DMF consists of Local Refinement Units (LRU) and Heterogenous Transmission Modules (HTM).
LRU is used to achieve feature alignment.
HTM includes a Shared Encoding Unit (SEU) and two Mutual Fusion Units (MFU).
The stacked HTM enables the reuse of deep features from CNNs and Transformers.

\textbf{Local Refinement Unit.}
When adopting different backbones, the dimensions of output features may not be consistent.
Thus, we first utilize the LRU to adjust the spatial sizes and channel dimensions of convolutional and transformeral features.
Besides, LRU is useful for reducing the computational burden and increasing the efficiency.
Meanwhile, for the transformeral features, LRU employs local interactions between addjacent patches, which is beneficial for introducing the spatial information.
Specifically, taking the transformeral features as an example, $\textbf{X}_t\ \in R^{D_t\times N}$ is firstly reshaped to $\hat{\textbf{F}}_t \in R^{D_t\times H_t\times W_t}$.
Then, the depth-wise convolution and point-wise convolution are utilized by:
\begin{equation}\label{3}
\textbf{F}  = \delta( \phi( \hat{\textbf{F}}_t  ) ),
\end{equation}
\begin{equation}\label{4}
\textbf{F}_t  = \delta( \varphi (\textbf{F} ) ),
\end{equation}
where $\delta$ presents the PReLU~\cite{he2015delving}.
$\phi$ and $\varphi$ present the depth-wise convolution and point-wise convolution with a Batch Normalization (BN) layer~\cite{ioffe2015batch}, respectively.
The depth-wise convolution is applied to change the spatial size of features, then the point-wise convolution to reduce the channel number.
Similarly, the convolutional features $\hat{\textbf{F}}_c \in R^{D_c \times H_c \times W_c}$ directly copied from $\textbf{X}_c$ is also passed into another LRU.
Finally, we obtain the refined features $\textbf{F}_c \in R^{D \times H \times W}$ and $\textbf{F}_t \in R^{D \times H \times W}$.
After applying the Generalized-Mean (GeM) pooling~\cite{radenovic2018fine} on $\textbf{F}_c$ and $\textbf{F}_t$, $f_{(c,0)}'$ and $f_{(t,0)}'$ are utilized to supervise the CNN and Transformer backbones, respectively.

\textbf{Shared Encoding Unit.}
In this work, SEU is designed to enable deep feature enhancement.
We employ a self-attention mechanism to explore the dependencies among features, highlighting informative regions and suppressing irrelevant noises.
As shown in Fig.~\ref{fig3}(b), the transformeral features $\textbf{F}_t$ obtained from LRU are first flattened.
Then it is concatenated with the global token $f_{(t,0)}'$ to form $\textbf{F}_{(t,0)}' \in R^{D\times (H_tW_t+1)}$. After that, $\textbf{F}_{(t,0)}'$ is passed into SEU.
In SEU, the Layer Normalization (LN)\cite{xiong2020layer}, Multi-Head Self-Attention (MHSA)\cite{dosovitskiy2020image} and Feed-Forward Network (FFN)\cite{dosovitskiy2020image} with residual connection \cite{he2016identity} are utilized to yield transformeral features:
\begin{equation}\label{5}
\hat{\textbf{S}}_{(t,0)} = \mathbf{MHSA}(\mathbf{LN} (\textbf{F}_{(t,0)}' ) ) + \textbf{F}_{(t,0)}',
\end{equation}
\begin{equation}\label{6}
\textbf{S}_{(t,0)} = \mathbf{FFN} ( \mathbf{LN}( \hat{\textbf{S}}_{(t,0)} ) ) + \hat{\textbf{S}}_{(t,0)},
\end{equation}
\begin{equation}
    \textbf{S}_{(t,0)} =  \mathbf{Concat}(\textbf{S}^{g}_{(t,0)}, \textbf{S}^{l}_{(t,0)}).
\end{equation}
Similarly, the convoluational features $\textbf{F}_{(c,0)}'$ is also passed into another SEU to obtain $\textbf{S}_{(c,0)}$ for deep feature enhancement.
The SEU is shared in each HTM and encodes the convolutional and transformeral features for MFU.

\textbf{Mutual Fusion Unit.}
MFU allows for global and local interactions between features of different attributes, forming a unified fusion framework through symmetric operations on tokenlized features.
Specifically, MFU utilizes the cross-attention mechanism to achieve the deep feature fusion between the convolutional and transformeral features.
As shown in Fig.~\ref{fig3}(c), we first feed the global convolutional feature $\textbf{S}_{(c,0)}^g$ and local transformeral features $\textbf{S}_{(t,0)}^l$ into MFU.
In MFU, we employ a Multi-Head Cross-Attention (MHCA), which consists of $N_h$ heads.
More specifically, in the $h$-th head of the CNN branch, the global token $\textbf{S}_{(c,0)}^g$ is passed into a linear transformation to generate $\textbf{Q}_{(c,0)}$, which is seen as the \emph{Query}.
The local features $\textbf{S}_{(t,0)}^l$ from the Transformer branch are passed into two linear transformations to generate $\textbf{K}_{(t,0)}$ and $\textbf{V}_{(t,0)}$, which are seen as the \emph{Key} and \emph{Value}, respectively.
Here, $\textbf{Q}_{(c,0)}$, $\textbf{K}_{(t,0)}$, $\textbf{V}_{(t,0)} \in R^{d\times H_tW_t}$, $d= \frac{D}{N_h}$.
By utilizing them, the cross-attention is defined as:
\begin{equation}\label{8}
\hat{\textbf{M}}^h_{(c,0)} = \sigma (\frac{\textbf{Q}_{(c,0)} \textbf{K}_{(t,0)}^ \top}{\sqrt{d}})\textbf{V}_{(t,0)},
\end{equation}
where $\sigma$ is the softmax activation function and $(\cdot)^\top$ means the transposition operation.
The outputs of multiple heads ($\hat{\textbf{M}}^1_{(c,0)},\cdots, \hat{\textbf{M}}^h_{(c,0)},\cdots, \hat{\textbf{M}}^{N_h}_{(c,0)}$) are concatenated to be $\hat{\textbf{M}}_{(c,0)}$.
In the $h$-th head of MHCA, the global feature $\textbf{S}_{(t,0)}^g$ from the Transformer branch is passed into a linear transformation to generate $\textbf{Q}_{(t,0)}$, which is seen as the  \emph{Query}.
The local features $\textbf{S}_{(c,0)}^l$ from the CNN branch are passed into two linear transformations to generate $\textbf{K}_{(c,0)}$ and $\textbf{V}_{(c,0)}$, which are seen as the \emph{Key} and \emph{Value}, respectively.
Similarly, $\textbf{Q}_{(t,0)}$, $\textbf{K}_{(c,0)}$, $\textbf{V}_{(c,0)} \in R^{d\times H_cW_c}$.
The interaction is formulated as:
\begin{equation}\label{8}
\hat{\textbf{M}}^h_{(t,0)} = \sigma (\frac{\textbf{Q}_{(t,0)} \textbf{K}_{(c,0)}^ \top}{\sqrt{d}})\textbf{V}_{(c,0)}.
\end{equation}
Here, ($\hat{\textbf{M}}^1_{(t,0)},\cdots, \hat{\textbf{M}}^h_{(t,0)},\cdots, \hat{\textbf{M}}^{N_h}_{(t,0)}$) are concatenated to be $\hat{\textbf{M}}_{(t,0)}$.
Then, by utilizing different FFNs, the features $\hat{\textbf{M}}_{(c,0)}$ and $\hat{\textbf{M}}_{(t,0)}$ are further encoded to obtain $f_{(c,1)}'$ and $f_{(t,1)}'$:
\begin{equation}\label{8}
f_{(c,1)}' = \mathbf{FFN}(\hat{\textbf{M}}_{(c,0)}) + \hat{\textbf{M}}_{(c,0)},
\end{equation}
\begin{equation}\label{8}
f_{(t,1)}' = \mathbf{FFN}(\hat{\textbf{M}}_{(t,0)}) + \hat{\textbf{M}}_{(t,0)}.
\end{equation}
Here, each FFN has two linear projections with the Gaussian Error Linear Unit (GELU)~\cite{hendrycks2016gaussian}.
Finally, by utilizing the cross-attention mechanism, our proposed MFU could fuse two types of deep features to generate the unified representations.

\textbf{Stacking of HTM.}
In this work, we combine a SEU and two MFUs as the Heterogenous Transmission Module (HTM).
As shown in Fig. \ref{fig2}, we continuously stack HTM to fuse the deep features from the two-branch framework.
Noted that, $f_{(c,1)}'$ will be concatenated with the patch tokens in $F_{(c,0)}'$ again,
As shown in Fig. \ref{fig3} (a), in the $k$-1 step, $f_{(c,k-1)}'$ has fused the information from the Transformer branch.
Then, it is concatenated with initial deep features from $\textbf{F}_{(c,0)}'$, following self-attention to model their internal relationships again.
By interacting with the local features, the transmission of heterogeneous information is achieved.
The symmetry operation will be applied to $f_{(t,k-1)}'$.
Through this continuously stacking, we unify two types of deep features.
Thus, the mutual guidance and heterogenous transmission are achieved.
In this way, it enables DMF to use different information at multiple levels.
Then, the output $f_{(c,L)}'$ and $f_{(t,L)}'$  of the last layer are obtained for loss supervision.
Finally, as shown in Fig. \ref{fig2}, six features are supervised during training.
During testing, we concatenate these six features to form $f_{a}$ for the final retrieval.
\subsection{Objective Function}
As shown in Fig.~\ref{fig2}, six output features from DFE and DMF are supervised by the label smoothing cross-entropy loss~\cite{szegedy2016rethinking} and triplet loss~\cite{hermans2017defense}.
More specifically, the features $\hat{f_c}$ and $\hat{f_t}$ from the CNN branch and Transformer branch in DFE are first supervised for dual-branch learning.
Taking $\hat{f_c}$ as an example, the label smoothing cross-entropy loss can be defined by:
\begin{equation}\label{15}
L_{ce} = y^j log \frac{exp(W_j \hat{f_c})}{\sum_{j}^{J} exp(W_j \hat{f_c})},
\end{equation}
where $y^j$ is the corresponding ground-truth label.
$J$ denotes the total number of classes.
In a mini-batch, we build a feature triplet set $\Omega = \{\emph{\textbf{F}}_a, \emph{\textbf{F}}_p, \emph{\textbf{F}}_n\}$.
Here, $\emph{\textbf{F}}_a$ presents the anchor features.
$\emph{\textbf{F}}_p$ and $\emph{\textbf{F}}_n$ present its positive features and negative features, respectively.
Thus, the triplet loss can be defined by:
\begin{equation}
\begin{split}
L_{tri} = \log \left[1+ \exp \left(\left\|\emph{\textbf{F}}_{a}-\emph{\textbf{F}}_{p}\right\|_{2}^{2}-\left\|\emph{\textbf{F}}_{a}-\emph{\textbf{F}}_{n}\right\|_{2}^{2}\right)\right],
\end{split}
\end{equation}
where $\|\cdot \|_2$ refers to the L2 norm.
Afterwards, to improve the discriminative ability~\cite{zhang2017amulet}, we also employ the supervision after aligned features ($f_{(c,0)}'$ and $f_{(t,0)}'$) and fused features ($f_{(c,L)}'$ and $f_{(t,L)}'$) in DMF.
Finally, we optimize the whole framework with six features simultaneously by a total loss:
\begin{equation}
\begin{split}
L_{total} = \sum_{g=1}^{6} {( L_{ce}^g + L_{tri}^g ).}
\end{split}
\end{equation}
%-------------------------------------------------------------------------
\section{Experiments}
\subsection{Datasets and Evaluation Metrics}
We conduct experiments on three large-scale ReID datasets, \emph{i.e.}, Market1501~\cite{zheng2015scalable}, DukeMTMC~\cite{dosovitskiy2020image} and MSMT17~\cite{wei2018person}.
Market1501 consists of 32,668 images of 1,501 identities.
DukeMTMC consists of 16,522 training images of 1,404 identities.
It has 2,228 queries images and 17,661 gallery images.
MSMT17 is the most challenging dataset.
There are 126,441 images of 4,101 identities observed under 15 different cameras.
All images provide their camera IDs in three datasets, which can be used as additional information when training and testing.
Following previous works, we adopt the mean Average Precision (mAP) and Cumulative Matching Characteristics (CMC) at Rank-1 as our evaluation metrics.
\subsection{Implementation Details}
In this work, we realize our model with the Pytorch toolbox, and conduct experiments on two NVIDIA A800 GPUs.
In DFE, we employ the pre-trained CNNs and Transformers on the ImageNet dataset~\cite{deng2009imagenet} as our backbones.
All person images are resized to 256$\times$128.
When training, these images are augmented by random horizontal flipping, random cropping and random erasing~\cite{zhong2020random}.
The min-batch size is set to 128, containing 8 identities and 16 images per identity.
During training, we adopt the Stochastic Gradient Descent (SGD) optimizer with a momentum of 0.9 and the weight decay of 1e$^{-4}$.
Besides, the learning rate is initialized as 5e$^{-4}$.
In the first 10 epoch, we adopt a warmup strategy with a linearly growing learning rate to 5e$^{-3}$.
Then, it decreases with a cosine decay for the next 170 epochs.
%
%-------------------------------------------------------------------------
\begin{table*}[htpb]
    \centering
    \caption{Quantitative comparison on Market1501, DukeMTMC and MSMT17. $\uparrow$ means the input image resolution is $384\times 128$, otherwise $256\times 128$. * means that overlapping is used in ViT. Pattern stands for the typical types of features present in the model. Conv means convolutional features while Trans means transformeral features. The best results are in bold, and the second best results are underlined.}
    \renewcommand\arraystretch{1.24}
    \setlength\tabcolsep{7pt}
    \begin{tabular}{lcccccccc}
        \Xhline{0.8pt}
        \multirow{2}{*}{Method}  & \multirow{2}{*}{Backbone}& \multirow{2}{*}{Pattern} & \multicolumn{2}{c}{Market1501} & \multicolumn{2}{c}{DukeMTMC} & \multicolumn{2}{c}{MSMT17} \\ \cline{4-9}
        &      &                   & mAP           & Rank-1        & mAP              & Rank-1        & mAP           & Rank-1           \\ \hline
        MGN ${\uparrow}$\cite{wang2018learning} & ResNet50& Conv & 86.9 & 95.7 & 78.4 & 88.7 & - & - \\
        BFE ${\uparrow}$\cite{dai2019batch}  & ResNet50& Conv & 86.2 & 95.3 & 75.9 & 88.9 & 51.5 & 78.8 \\
        OSNet\cite{zhou2019omni}  & OSNet& Conv & 84.9 & 94.8 & 73.5 & 88.6 & 52.9 & 78.7 \\
        HOReID\cite{wang2020high}  & ResNet50& Conv & 84.9 & 94.2 & 75.6 & 86.9 & - & - \\
        CDNet\cite{li2021combined}  & CDNet& Conv & 86.0 & 95.1 & 76.8 & 88.6 & 54.7 & 78.9 \\
        TransReID\cite{he2021transreid} & ViT-B/16$^*$& Trans & 88.9 & 95.2 & 82.0 & 90.7 & 67.4        & 85.3         \\
        PFD\cite{wang2022pose}  & ViT-B/16$^*$& Trans & 89.7 & 95.5 & - & - & 64.4 & 83.8 \\
        AAformer\cite{zhu2021aaformer}  & ViT-B/16$^*$& Trans & 87.7 & 95.4 & 80.0 & 90.1 & 62.6 & 83.1 \\
        GLTrans\cite{wang2024other} & ViT-B/16 & Trans & 90.0 & 95.6 & 82.4 & 90.7 & 69.0 & 85.8 \\
        ALDER ${\uparrow}$\cite{zhang2021seeing} & ResNet50& Conv + Trans & 88.9 & 95.6 & 78.9 & 89.9 & 59.1       & 82.5         \\
        IGOAS ${\uparrow}$\cite{zhao2021incremental}  & ResNet50& Conv + Trans & 84.1 & 93.4 & 75.1 & 86.9          & -           & -            \\
        HAT\cite{zhang2021hat}  & ResNet50& Conv + Trans & 89.5 & 95.6 & 81.4 & 90.4 & 61.2 & 82.3 \\
        RGA-SC\cite{zhang2020relation}  & ResNet50& Conv + Trans & 88.4 & 96.1 & - & - & 57.5        & 80.3         \\
        ABDNet ${\uparrow}$\cite{chen2019abd}  & ResNet50& Conv + Trans & 88.3 & 95.6 & 78.6 & 89.0 & 60.8        & 82.3         \\
        FED\cite{wang2022feature}  & ViT-B/16& Conv + Trans & 86.3 & 95.0 & - & - & - & - \\
        NFormer\cite{wang2022nformer}  & ResNet50& Conv + Trans& 91.1 & 94.7 & \underline{83.5}            & 89.4             & 59.8           & 77.3            \\\hline
        FusionReID       & ResNet50 + ViT-B/16*   & Conv + Trans & \textbf{91.7} & \textbf{96.3} & \underline{83.5} & \textbf{91.0}          & \underline{69.5}        & \underline{86.7}         \\
        {FusionReID}       &{ResNet50 + ViT-B/16}    & {Conv + Trans} & {90.9} & {95.9} & {82.9} & {90.3}          & {68.3}        & {86.1}         \\
        {FusionReID ${\uparrow}$}       & {ResNet50 + ViT-B/16*}    & {Conv + Trans} & {\underline{91.6}} & {\underline{96.2}} & {\textbf{83.7}} & {\textbf{91.0}}          & {\textbf{70.5}}        & {\textbf{87.3}}         \\
        {FusionReID ${\uparrow}$}       & {ResNet50 + ViT-B/16}    & {Conv + Trans} & {91.1} & {96.0} & {\underline{83.5}} & {\underline{90.8}}          & {\underline{69.5}}        & {{86.6}}  \\    \Xhline{0.8pt}
    \end{tabular}
    \label{tab1}
\end{table*}
%---------------------------------------------------------------------
\subsection{Comparison with State-of-the-arts}
Tab.~\ref{tab1} shows the evaluation on three public ReID benchmarks, \emph{i.e.}, Market1501, DukeMTMC and MSMT17.

\textbf{Market1501}:
On Market1501, we can see that our method achieves the best mAP and Rank-1.
{Noted that, RGA-SC~\cite{zhang2020relation} gains a comparable Rank-1 on Market1501, which learns a global attention to refine the features.}
Different from RGA-SC, we utilize HTM to encode and fuse two types of deep features.
Thus, our method has a higher mAP than RGA-SC.
The best performances show that our proposed framework is able to unify the strengths of CNNs and Transformers to get more discriminative representations.

\textbf{DukeMTMC}:
Our method achieves the best Rank-1 and mAP on DukeMTMC.
Noted that, NFormer~\cite{wang2022nformer} utilizes Transformers to model the relationships between all images, obtaining robust representations.
Different from NFormer, we intergrate the heterogeneous information from CNNs and Transformers, resulting in a higher Rank-1 and mAP.

\textbf{MSMT17}:
As shown in Tab.~\ref{tab1}, our method achieves the best mAP and Rank-1 on MSMT17.
It is noteworthy that although NFormer performs well on small datasets, it performs poorly on the most challenging large-scale dataset, MSMT17.
Although it has a lower number of parameters and FLOPs, our method outperforms it by about 10\% on mAP and Rank-1.
The results also demonstrate the stronger generalization performance of our method on different scale datasets.
Overall, our method attains better performances on three public ReID benchmarks than most state-of-the-art methods.
The best mAP and Rank-1 demonstrate the superiors of our method even with low image resolutions, especially on the large-scale dataset.

{\textbf{Performances with Different Configurations}:
In the last four rows of Tab.~\ref{tab1}, we compare the performance of FusionReID with different configurations, including the input image resolution and the overlapping of ViT.
Generally, using high-resolution images and patch overlapping in ViT can improve the performance of FusioReID on all three datasets.
High-resolution images contain more visual information, and the patch overlapping enhances correlations between neighboring areas.
These superior performances demonstrate the generalization ability of our proposed FusionReID.}
\subsection{Ablation Study}
\textbf{Effectiveness of Key Components.}
We conduct experiments to verify the effectiveness of proposed key components.
The ablation results are shown in Tab.~\ref{tab2}.
Comparing the results of the first three rows, one can see that although DFE only concatenates two types of backbone features, higher ReID accuracies are achieved.
Furthermore, we employ our DMF after DFE.
The results are reported in the last four rows of Tab.~\ref{tab1}.
When utilizing LRU, the performance has been slightly improved.
After feature alignment in LRU, we further utilize SEU or MFU to encode and fuse deep features.
Therefore, compared with DFE, the utilization of SEU gains 1.0\% mAP and 0.7\% Rank-1 improvements, and the utilization of MFU gains 1.2\% mAP and 0.6\% Rank-1 improvements.
Furthermore, our method combines all proposed key components and achieves the best performance, \emph{i.e.}, 69.5\% mAP and 86.7\% Rank-1.
Overall, the gradual increase in performances validates the effectiveness of our proposed key components.
\begin{table}[t]
    \renewcommand\arraystretch{1.24}
    \setlength\tabcolsep{3.0pt}
    \caption{Ablation study of key components on MSMT17. ResNet refers to ResNet50. ViT-B refers to ViT-B/16$_{s=12}$.}
    \label{tab2}
    \centering
\begin{tabular}{c|c|ccc|cc|cc}
\hline
\multirow{2}{*}{Method} & \multicolumn{1}{c|}{\multirow{2}{*}{DFE}} & \multicolumn{3}{c|}{DMF} & \multicolumn{2}{c|}{MSMT17} & Params & FLOPs \\ \cline{3-7}
                        & \multicolumn{1}{c|}{}                     & LRU    & SEU    & MFU    & mAP          & Rank-1           & (M)                           & (G)                         \\ \hline
ResNet               & \ding{53}                         &\ding{53}      &\ding{53}      &\ding{53}        & 54.3         & 78.1         & 23.5                      & 4.1                      \\
ViT-B               &\ding{53}                          &\ding{53}        &\ding{53}        &\ding{53}        & 64.6         & 83.2         & 85.7                      & 18.1                     \\ \hline
Method1                    &\ding{51}                          & \ding{53}       & \ding{53}       &  \ding{53}        & 67.5         & 85.1         & 109.1                     & 22.1                     \\
Method2                  & \ding{51}                         &  \ding{51}      & \ding{53}       & \ding{53}       & 67.7         & 85.4         & 111.3                     & 22.4                     \\
Method3                  & \ding{51}                         & \ding{51}       & \ding{51}       & \ding{53}        & 68.7         & 86.1         & 125.5                     & 27.3                     \\
Method4                   & \ding{51}                         & \ding{51}       & \ding{53}       & \ding{51}       & 68.9         & 86.0         & 139.7                     & 23.3                     \\
Method5             & \ding{51}                         & \ding{51}       & \ding{51}       & \ding{51}       & 69.5         & 86.7         & 153.8                     & 28.1                     \\ \hline
\end{tabular}
\vspace{-2mm}
\end{table}
%-----------------------------------------------------
\begin{figure}[t]
    \centering
    \includegraphics[width=0.45\textwidth]{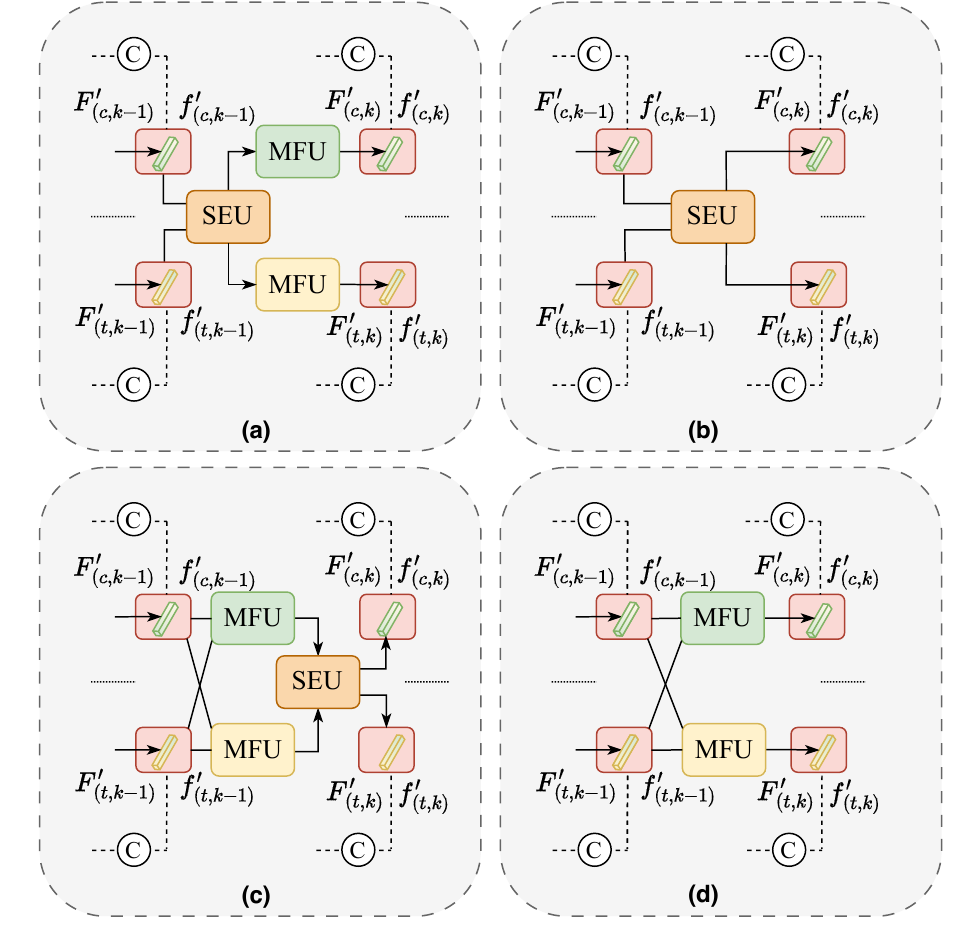}
    \vspace{-2mm}
    \caption{Various structures in HTM. (a) SEUs in front and MFUs in back; (b) Only shared SEUs are continuously stacked; (c) MFUs in front and SEU in back; (d) Only unshared MFUs are stacked.}
    \label{fig5}
    \vspace{-2mm}
    \end{figure}
%-----------------------------------------------------
\begin{table}[t]    \renewcommand\arraystretch{1.24}
    \caption{Ablation study of various structures in HTM. ``Front'' means in the front of HTM, while ``Back'' means in the back of HTM.}
    \label{tab3}
    \centering
    \begin{tabular}{cc|cc|cc}
        \hline
        \multicolumn{2}{c|}{Method} & \multicolumn{2}{c|}{MSMT17} &Params &FLOPs \\
        SEU   & MFU   & mAP  & Rank-1 & (M)  & (G)   \\ \hline
        Front & Back  & 69.5 & 86.7  & 21.3 & 2.9\\
        Only  & -     & 68.7 & 86.1 & 7.1 & 2.5\\
        Back  & Front & 68.5 & 86.2 & 21.3 & 2.9\\
        -     & Only  & 68.9 & 86.0 & 14.2 & 0.5\\\hline
    \end{tabular}
    \vspace{-2mm}
\end{table}
%-----------------------------------------------------
\begin{table}[t]    \renewcommand\arraystretch{1.24}
\caption{{Ablation study of utilizing shared weights in SEU and MFU.}}
\label{tab4}
\centering
\begin{tabular}{cc|cc|cc}
    \hline
    \multicolumn{2}{c|}{Method} & \multicolumn{2}{c|}{MSMT17} &Params &FLOPs \\
    SEU   & MFU   & mAP  & Rank-1 & (M)  & (G)   \\ \hline
    Shared & Shared  & 69.1 & 86.6  & 14.2 & 2.9\\
    Shared & Unshared  & 69.5 & 86.7 & 21.3 & 2.9\\
    Unshared  & Shared & 68.7 & 86.0 & 21.3 & 2.9\\
    Unshared  & Unshared  & 69.2 & 86.3 & 28.4 & 2.9\\\hline
\end{tabular}
\vspace{-2mm}
\end{table}
%-----------------------------------------------------
\begin{table}[t]
    \renewcommand\arraystretch{1.24}
    \caption{{Ablation study of different pooling structures in DMF.}}
    \label{tab6}
    \centering
    \begin{tabular}{c|cc|cc}
        \hline
        \multirow{2}{*}{Method} & \multicolumn{2}{c|}{MSMT17} & \multirow{1}{*}{Params} & \multirow{1}{*}{FLOPs} \\
                                & mAP  & Rank-1                &  (M)                           & (G)                          \\\hline
        With GeMP        & 69.5 & 86.7                  & 153.8 &28.1                                  \\
        With GAP & 68.6 & 86.2               & 153.8 &28.1                                 \\\hline
    \end{tabular}
    \vspace{-2mm}
\end{table}
%-----------------------------------------------------
\begin{table}[t]
    \centering
	\renewcommand\arraystretch{1.24}
    \setlength\tabcolsep{3.4pt}
	\caption{Ablation study of different backbones on MSMT17.}
	\label{tab5}
	\begin{tabular}{cc|ccc|cc}
		\hline
		\multicolumn{2}{c|}{Method} & \multicolumn{3}{c|}{MSMT17}& Params & FLOPs \\ \cline{1-5}
		\multicolumn{1}{c|}{CNN}                        & Transformer                & Feature & mAP  & Rank-1  & (M)       & (G) \\ \hline
		\multicolumn{1}{c|}{ResNet50}                   & -                          & $f$     & 54.3 & 78.1 & 23.5& 4.1\\
		\multicolumn{1}{c|}{ResNet101}                  & -                          & $f$     & 58.8 & 81.4 & 42.5& 6.5\\
		\multicolumn{1}{c|}{ResNet152}                  & -                          & $f$     & 61.0 & 82.7 & 58.1& 8.9\\ \hline
        \multicolumn{1}{c|}{-}                          & T2T-ViT-14               & $f$     & 50.9 & 73.7 & 21.0 & 2.8\\
		\multicolumn{1}{c|}{-}                          & ViT-B/16$^*$               & $f$     & 64.6 & 83.2 & 85.7 & 18.1  \\
		\multicolumn{1}{c|}{-}                          & DeiT-B/16$^*$              & $f$     & 63.3 & 83.1 & 85.7 & 18.1  \\
        % \multicolumn{1}{c|}{-}                          & ResTv2-L                 & $f$     & - & - \\
        \hline
		\multicolumn{1}{c|}{\multirow{9}{*}{ResNet50}}  & \multirow{3}{*}{T2T-ViT-14}  & $f_{(c,L)}'$   &  60.8  &  80.2&\multirow{3}{*}{89.0}&\multirow{3}{*}{11.4} \\
		\multicolumn{1}{c|}{}                           &                            & $f_{(t,L)}'$ & 56.7 & 78.3 \\
		\multicolumn{1}{c|}{}                           &                            & $f_{a}$ & 61.9 & 81.6 \\ \cline{2-7}
		\multicolumn{1}{c|}{}                           & \multirow{3}{*}{ViT-B/16$^*$}  & $f_{(c,L)}'$   &  66.0  &  85.4&\multirow{3}{*}{153.8}&\multirow{3}{*}{28.1} \\
		\multicolumn{1}{c|}{}                           &                            & $f_{(t,L)}'$ & 67.8 & 85.6 \\
		\multicolumn{1}{c|}{}                           &                            & $f_{a}$ & 69.5 & 86.7 \\ \cline{2-7}
		\multicolumn{1}{c|}{}                           & \multirow{3}{*}{DeiT-B/16$^*$} & $f_{(c,L)}'$ & 64.9 & 84.8&\multirow{3}{*}{153.8}&\multirow{3}{*}{28.1} \\
		\multicolumn{1}{c|}{}                           &                            & $f_{(t,L)}'$ & 67.0 & 85.4 \\
		\multicolumn{1}{c|}{}                           &                            & $f_{a}$ & 69.7 & 86.4 \\ \hline
		\multicolumn{1}{c|}{\multirow{9}{*}{ResNet101}} & \multirow{3}{*}{T2T-ViT-14}  & $f_{(c,L)}'$   &  62.7  &  83.0&\multirow{3}{*}{107.0}&\multirow{3}{*}{13.9} \\
		\multicolumn{1}{c|}{}                           &                            & $f_{(t,L)}'$ & 58.4 & 79.9 \\
		\multicolumn{1}{c|}{}                           &                            & $f_{a}$ & 63.4 & 82.6 \\ \cline{2-7}
		\multicolumn{1}{c|}{}                           & \multirow{3}{*}{ViT-B/16$^*$}  & $f_{(c,L)}'$   & 66.9   & 85.8&\multirow{3}{*}{172.8}&\multirow{3}{*}{30.5}  \\
		\multicolumn{1}{c|}{}                           &                            & $f_{(t,L)}'$ & 68.4 & 86.1 \\
		\multicolumn{1}{c|}{}                           &                            & $f_{a}$ & 70.5 & 87.3 \\ \cline{2-7}
		\multicolumn{1}{c|}{}                           & \multirow{3}{*}{DeiT-B/16$^*$} & $f_{(c,L)}'$ & 66.0 & 85.7&\multirow{3}{*}{172.8}&\multirow{3}{*}{30.5} \\
		\multicolumn{1}{c|}{}                           &                            & $f_{(t,L)}'$ & 67.5 & 85.8 \\
		\multicolumn{1}{c|}{}                           &                            & $f_{a}$ & 70.3 & 87.1 \\ \hline
		\multicolumn{1}{c|}{\multirow{9}{*}{ResNet152}} & \multirow{3}{*}{T2T-ViT-14}  & $f_{(c,L)}'$   &  63.9  &  83.9&\multirow{3}{*}{123.6}&\multirow{3}{*}{16.3} \\
		\multicolumn{1}{c|}{}                           &                            & $f_{(t,L)}'$ & 59.1 & 80.2 \\
		\multicolumn{1}{c|}{}                           &                            & $f_{a}$ & 64.5 & 83.4 \\ \cline{2-7}
		\multicolumn{1}{c|}{}                           & \multirow{3}{*}{ViT-B/16$^*$}  & $f_{(c,L)}'$   & 68.2   & 86.5&\multirow{3}{*}{188.5}&\multirow{3}{*}{32.9}  \\
		\multicolumn{1}{c|}{}                           &                            & $f_{(t,L)}'$ & 69.6 & 87.0 \\
		\multicolumn{1}{c|}{}                           &                            & $f_{a}$ & \textbf{71.8} & \textbf{87.9} \\ \cline{2-7}
		\multicolumn{1}{c|}{}                           & \multirow{3}{*}{DeiT-B/16$^*$} & $f_{(c,L)}'$ & 66.6 & 85.7&\multirow{3}{*}{188.5}&\multirow{3}{*}{32.9}  \\
		\multicolumn{1}{c|}{}                           &                            & $f_{(t,L)}'$ & 67.6 & 86.2 \\
		\multicolumn{1}{c|}{}                           &                            & $f_{a}$ & 70.9 & 87.5 \\ \hline
	\end{tabular}
	\label{tab9}
\vspace{-3mm}
\end{table}

\textbf{Effects of Various Structures in HTM.}
As shown in Fig. \ref{fig5}, we design various structures in HTM.
The ablation results with various structures are shown in Tab. \ref{tab3}.
Here, the computation cost only considers the HTM part.
One can find that the best performance is achieved by SEU first and then MFUs.
The reasons may be twofolds:
(1) The shared SEU in front can constrain the distributions of different deep features, which is helpful for the subsequent fusion.
(2) The unshared MFU in back can utilize deep mutual fusion to generate better features, which highlights the importance of feature fusion in the last step.
Among the four structures, the worst performance is achieved by MFU first and then SEU.
There are two reasons for the poor results:
(1) For the first HTM layer, the differences of deep features are significant, and it is difficult to fully integrate them through MFU.
(2) For the subsequent HTMs, they only change the weighted parameters of global features. When it goes directly to the MFU of next HTM, it is still fusing with each other’s initial deep features. The difference problem in feature distribution still exists.
Therefore, the order of SEU and MFU is very important.
Thus, we deploy SEU first and then MFU to achieve better deep fusions.

\textbf{Effects of Weight-Sharing in SEU and MFU.}
In Tab.~\ref{tab4}, we analyze the effect of weight-sharing in SEU and MFU.
The weight-sharing in SEU reduces feature differences across branches, improving MFU fusion.
Conversely, unshared MFUs allow each branch to focus on distinct fusion patterns, enhancing their effectiveness.
The combination of weight-sharing SEU and unshared MFU delivers the best overall performance.

\textbf{Effects of Different Pooling Structures in DMF.}
In Tab.~\ref{tab6}, we compare the performance of different pooling structures in DMF.
The GeMP clearly outperforms the Global Average Pooling (GAP), demonstrating its effectiveness to capture the distinctiveness with negligible computational cost.

\textbf{Complementarity between CNNs and Transformers.}
We employ different backbone networks in DFE to verify the complementarity between CNNs and Transformers.
In the CNN branch, we choose residual networks~\cite{he2016deep}, including ResNet50/101/152.
In the Transformer branch, we choose T2T-ViT-14~\cite{yuan2021tokens}, ViT-B/16~\cite{dosovitskiy2020image} and DeiT-B/16~\cite{touvron2021training}.
As shown in Tab.~\ref{tab5}, we find that the performance of ViT-B/16$^*$ and DeiT-B/16$^*$ outperforms ResNet50/101/152.
And for T2T-ViT-14, its corresponding features are also improved considerably with the help of CNNs.
When unifying ResNet50 and ViT-B/16$^*$, the fused feature $f_{(c,L)}'$ attains 66.0\% mAP and 85.4\% Rank-1, which has 11.7\% mAP and 7.3\% Rank-1 improvement over ResNet50.
Besides, the fused feature $f_{(t,L)}'$ attains 3.2\% mAP and 1.6\% Rank-1 improvement over the results of ViT-B/16$^*$, respectively.
Moreover, the concatenated feature $f_a$ has a better mAP and Rank-1.
These improvements suggest that our proposed framework has a significant effectiveness on unifying convolutional and transformeral features.
In the same way, when combining ResNet50/101/152 and DeiT-B/16$^*$, all the metrics are increased significantly.
In particular, when unifying ResNet152 and ViT-B/16$^*$, the overall feature $f_a$ obtains 71.8\% mAP and 87.9\% Rank-1.
The results clearly demonstrate the complementarity between CNNs and Transformers.
Meanwhile, they illustrates the generalization of our proposed method with different backbone networks.
\begin{figure}[t]
    \centering
\includegraphics[width=0.48\textwidth]{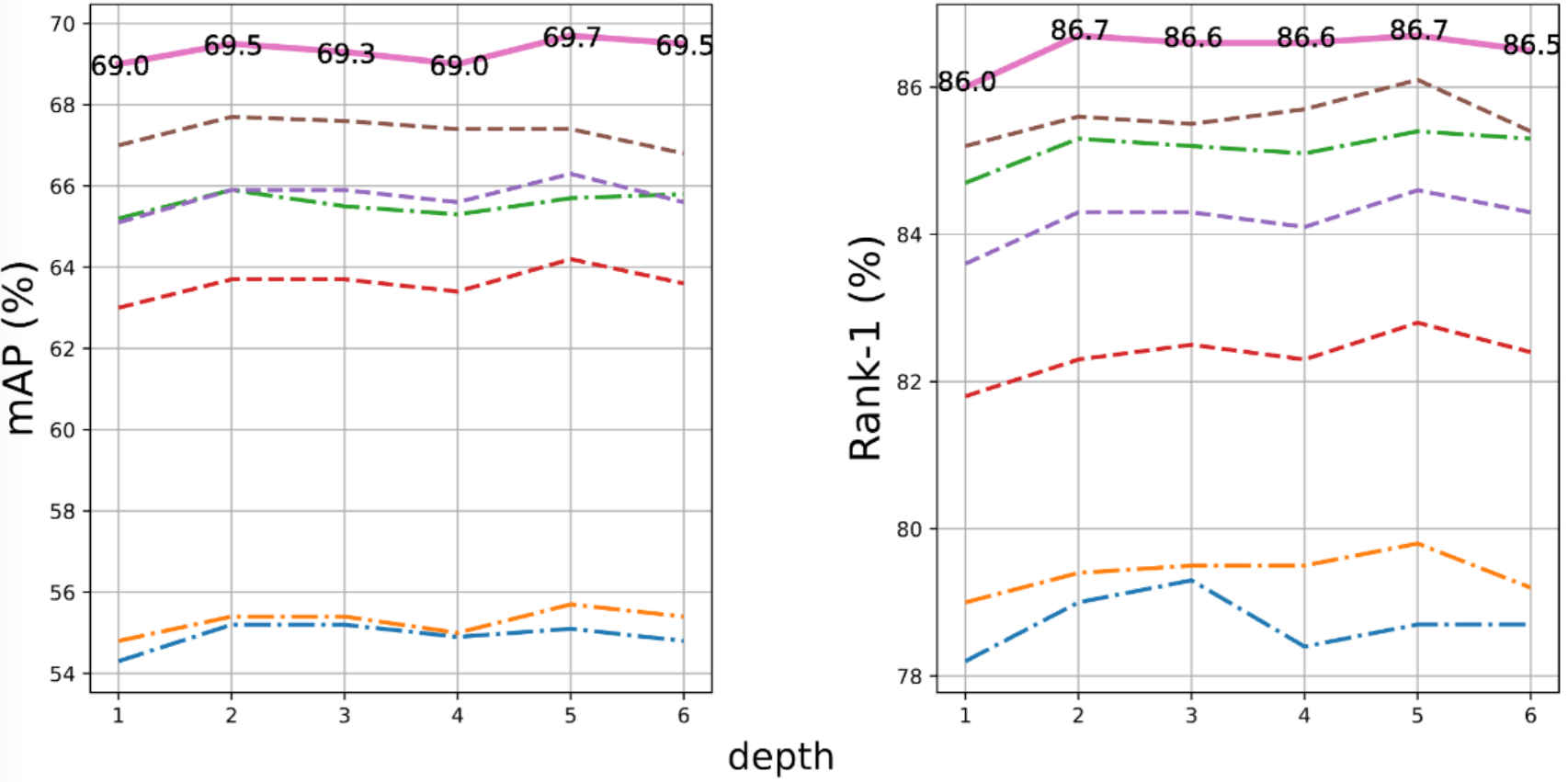}
\vspace{-2mm}
    \caption{Effects of stacked HTM on MSMT17.}
    \label{fig4}
    \vspace{-2mm}
\end{figure}

\textbf{Influence of the Stacked HTM.}
We conduct experiments on MSMT17 to verify the influence of the stacked HTM.
The results are shown in Fig.~\ref{fig4}.
One can see that the accuracy curve changes but not significantly.
As the number of stacked layers increases, the model performance gradually improves, especially from 1 to 2 layers.
The refined features (middle curves) after LRU are stronger than the original features (lower curves), while the fused features (upper curves) are better than the refined features after LRU.
The improvement indicates that the stacked HTM could better extract discriminative features.
However, the performance is saturated when stacked layers are more than 2.
Thus, we set the layers to 2 by default.

\begin{figure}[t]
    \centering
    \includegraphics[width=0.48\textwidth]{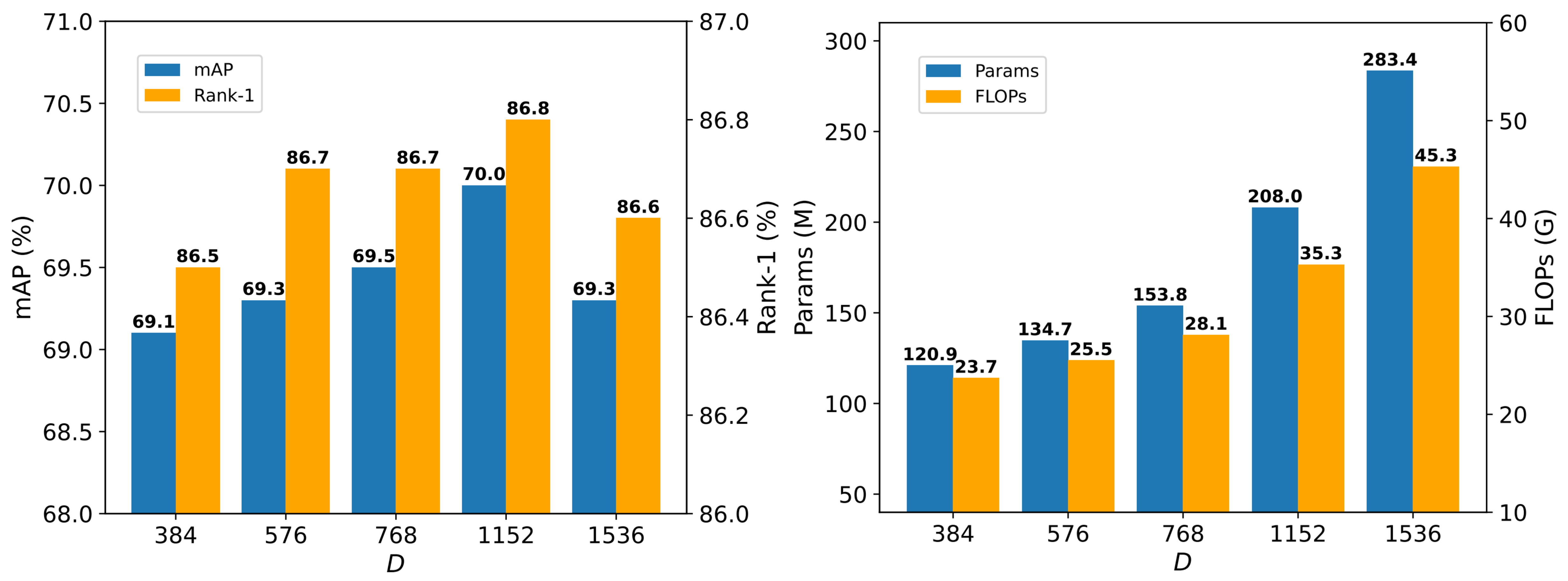}
\vspace{-2mm}
    \caption{Effects of feature dimension $D$ on MSMT17.}
    \label{fig6}
    \vspace{-4mm}
\end{figure}

\textbf{Influence of Fused Dimensions in DMF.}
Fig.~\ref{fig6} shows the influence of using different fused dimensions.
We change the feature dimensions from 384 to 1536.
We find that the performance is worst when the dimension is 384, and the performance is best when the dimension is 1152, achieving 70.0\% mAP and  86.8\% Rank-1.
With the dimensions keep rising, the performance of mAP and Rank-1 begins to decrease.
The reason may be that the model tends to over-fitting when the dimension is too high.
Although the fused dimension of 384 has the smallest number of model parameters and FLOPs, it outperforms most sate-of-the-art methods.
To avoid excessive model complexity, we set 768 as the default dimension.

\textbf{Analysis of Computational Cost.}
The ablation analysis of computational cost is also presented.
In Tab.~\ref{tab2}, as the accuracy increases, the computational cost increases.
In fact, the computational complexity of our proposed method is acceptable.
As shown in Fig.~\ref{fig6}, when we reduce the fused dimension to 384, the slight increase in model memory and computational complexity results in considerable accuracy gains.
For example, in comparison to TransReID, our approach demonstrates superior performance across three datasets, even with a moderate increase in parameters and FLOPs.
This also proves that the effectiveness of our method does not come from simple model stacking and parameter increasing.
%------------------------------------------------
%------------------------------------------------
\begin{figure*}[t]
    \centering
    \includegraphics[width=0.84\textwidth]{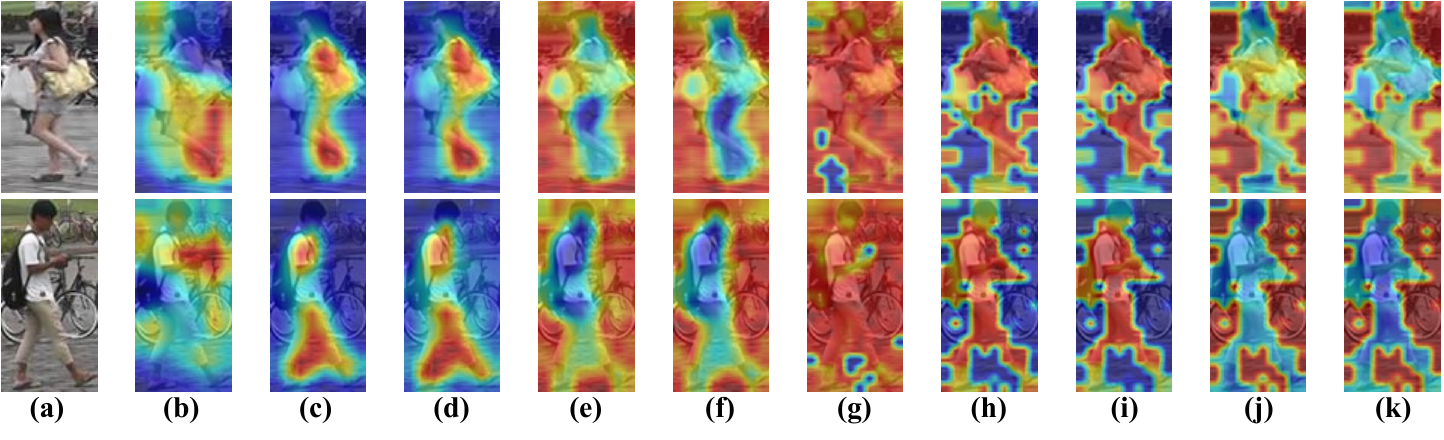}
\vspace{-2mm}
    \caption{Visualization of different deep features with Grad-CAM. (a) Original images; Convolutional features (b) from ResNet50 only, (c) before LRU, (d) after LRU, (e) after the first HTM, (f) after the second HTM; Transformeral features (g) from ViT-B/16$^*$ only, (h) before LRU, (i) after LRU, (j) after the first HTM, (k) after the second HTM.}
    \label{fig7}
    \vspace{-2mm}
\end{figure*}
%------------------------------------------------
\begin{figure*}[t]
    \centering
    \includegraphics[width=0.84\textwidth]{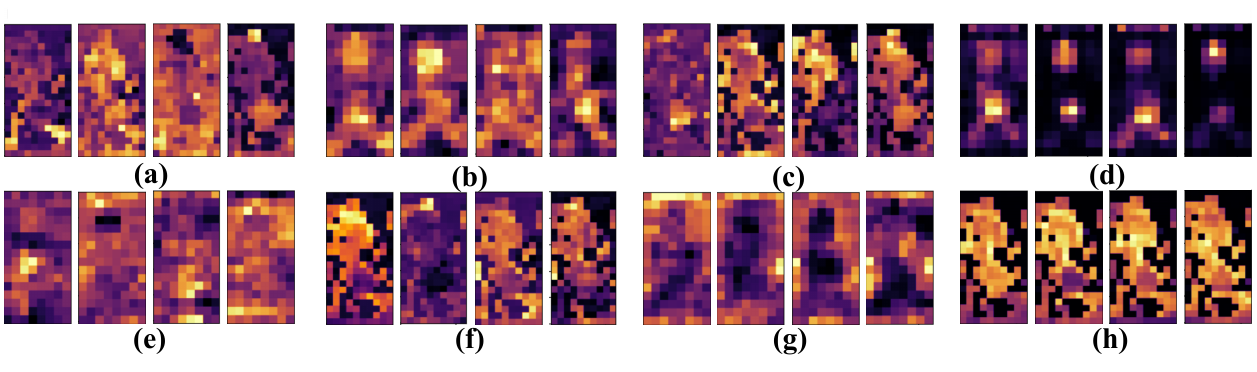}
\vspace{-2mm}
    \caption{{Visualization of attention weights in different heads. The first layer of (a) SEU and (b) MFU in the Transformer branch; The second layer of (c) SEU and (d) MFU in the Transformer branch; (e) to (h) represent
    the corresponding attention weights in the CNN branch.}}
    \label{fig9}
    \vspace{-2mm}
\end{figure*}
%------------------------------------------------
\begin{figure}[t]
    \centering
    \includegraphics[width=0.34\textwidth]{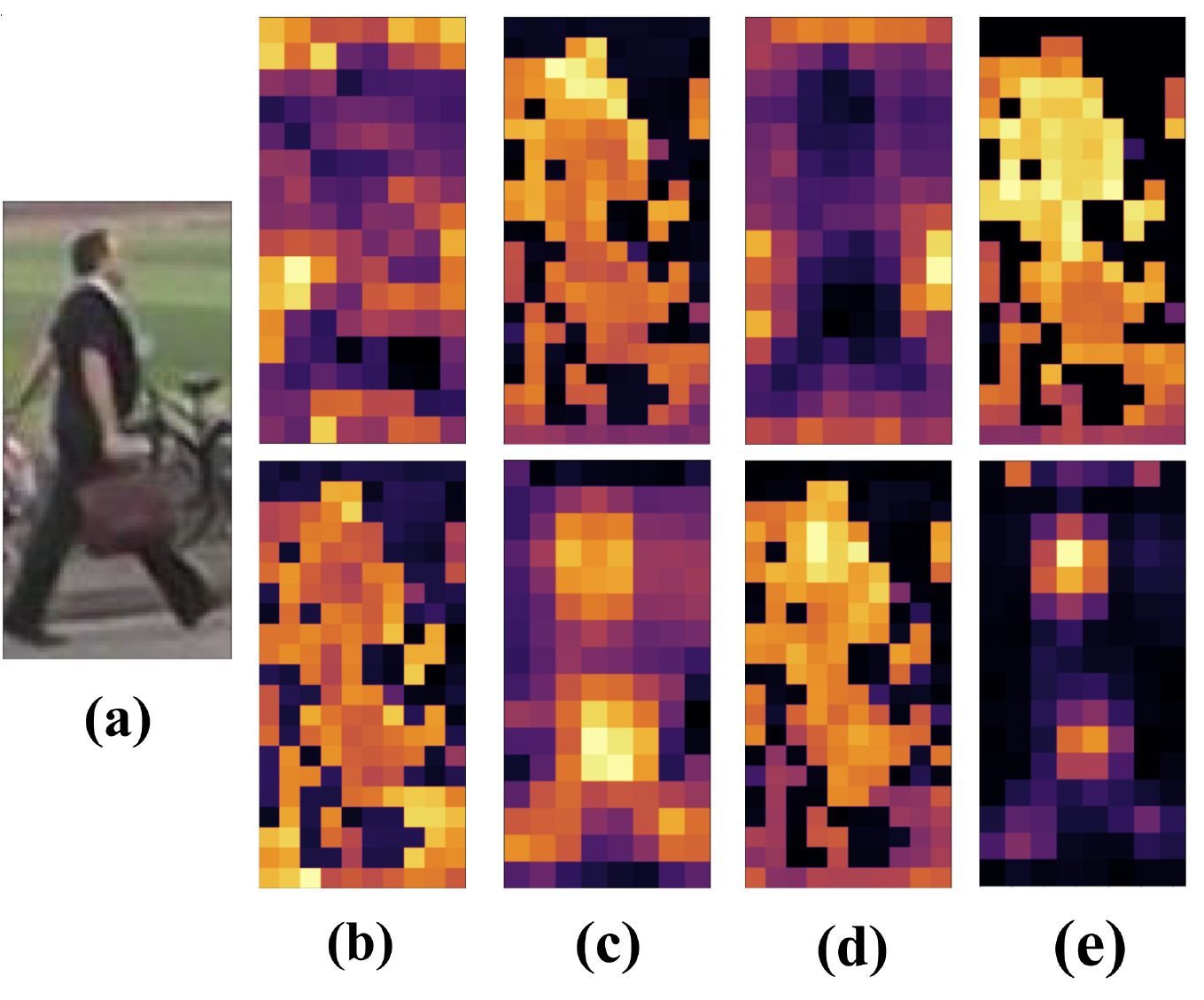}
    \vspace{-2mm}
    \caption{Attention weights in SEU and MFU. Top and bottom weights are from the CNN and Transformer branch, respectively. (a) The original image. Weights from (b) to (e) are in the first and second SEU and MFU, respectively.}
    \label{fig8}
    \vspace{-2mm}
    \end{figure}
%------------------------------------------------
\subsection{Visualization Analysis}
In this subsection, we further explain the effectiveness of our approach through visualization analysis.
First, we compare feature maps from typical CNNs and Transformers to verify their complementarities.
Besides, we visualize attention maps in our DMF to illustrate the effects of deep fusions.

\textbf{Visualization of Features in LRU and HTM.}
As shown in Fig.~\ref{fig7}, we utilize Grad-CAM~\cite{selvaraju2017grad} to visualize the feature maps in LRU and HTM.
Fig.~\ref{fig7}(a) shows the original images of six persons.
Comparing Fig.~\ref{fig7}(b) and Fig.~\ref{fig7}(g), one can see that convolutional features generally focus on meaningful local regions of persons, while transformeral features receives the global perception of persons.
The differences verify their unique strengthes between CNNs and Transformers.
Notably, Fig.~\ref{fig7}(g) and Fig.~\ref{fig7}(h) demonstrate that Transformers imitate the local perception of CNNs.
More edge information is captured by the transformeral features.
Comparing Fig.~\ref{fig7}(c), Fig.~\ref{fig7}(d), Fig.~\ref{fig7}(h) and Fig.~\ref{fig7}(i), one can see that the convolutional features after LRU appropriately highlight the regions of interest.
Different from Fig.~\ref{fig7}(d), Fig.~\ref{fig7}(e) shows the convolutional features after HTM begin to focus on more regions for comprehensive perceptions.
A similar phenomenon occurs in Fig.~\ref{fig7}(i) and Fig.~\ref{fig7}(j).
Noted that, the highlighted regions from CNNs and Transformers are complementary.
These visualizations provide a more intuitive understanding of the mutual fusion in our proposed methods.

\textbf{Visualization of Different Heads in HTM.}
In Fig.~\ref{fig9}, we visualize the attention weights of different heads in HTM, using the same example as in Fig.~\ref{fig8}.
As shown in Fig.~\ref{fig9}(a), different heads focus on different areas.
After the first MFU layer (Fig.~\ref{fig9}(b)), the Transformer branch shifts to key parts of the body, with a smoother weight distribution, similar to CNN.
After SEU (Fig.~\ref{fig9}(c)), attention regions change, and the second MFU layer (Fig.~\ref{fig9}(d)) narrows the focus to finer details like the head, legs, and briefcase.
Meanwhile, after the second MFU layer (Fig.~\ref{fig9}(h)), the CNN branch captures the entire human body with discriminative information.
These visualizations confirm the effectiveness of the mutual guidance in our proposed methods.

\textbf{Visualization of Attention Weights in SEU and MFU.}
As shown in Fig.~\ref{fig8}, we visualize the attention weights in SEU and MFU.
From the results, one can see that SEU can enhance their unique characteristics in the CNN branch and Transformer branch.
For example, in the CNN branch, SEU merely assigns high attention weights to the meaningful local regions.
In the Transformer branch, highlighted regions cover the human body and background with a global perception.
Meanwhile, Fig.~\ref{fig8}(c) and Fig.~\ref{fig8}(e) clearly show that the MFU fuses their unique characteristics, which are different with Fig.~\ref{fig8}(b) and Fig.~\ref{fig8}(d).
These visualizations could explain the mechanisms of feature enhancement and mutual fusion in SEU and MFU.
Moreover, the changes of attention weights in SEU and MFU validate the effectiveness of our DMF.
\section{Conclusion and Future Work}
In this paper, we propose a flexible fusion framework named FusionReID for image-based person ReID.
It comprises a Dual-branch Feature Extraction (DFE) and a Dual-attention Mutual Fusion (DMF).
In DFE, we employ CNNs and Transformers to extract deep features from a single image.
Besides, DMF consists of the Local Refinement Unit (LRU) and Heterogenous Transmission Module (HTM).
Through the continuous stacking of HTM, we unify heterogenous deep features from CNNs and Transformers.
Experiments on three large-scale ReID benchmarks demonstrate that our method attains superior performances than most state-of-the-arts.
Since the computation is still high, in the future, we will explore more lightweight fusion methods for the framework.
%%%------------------------------------------------------------------
\ifCLASSOPTIONcaptionsoff
  \newpage
\fi
\bibliographystyle{IEEEtran}
\bibliography{IEEEabrv,refs}
\end{document}